\documentclass[11pt]{article}
\usepackage{coling2020}
\usepackage{times}
\usepackage{url}
\usepackage{latexsym}

\usepackage{subcaption}
\usepackage{multicol}
\usepackage{multirow}
\usepackage{amssymb}
\usepackage{bm}
\usepackage{booktabs}
\usepackage{graphicx}
\usepackage{amsmath}

\newcommand{\citep}{\cite}
\newcommand{\citet}{\newcite}

\colingfinalcopy 

\newcommand{\ours}{\textsc{MoEBQA}}

\title{Mixture of Experts for Biomedical Question Answering}

\author{
    Damai Dai$^{\dag\ddag}$\thanks{~~Joint work of Peking University and Baidu Inc.},
    ~~Wenbin Jiang$^\ddag$, 
    ~~Jiyuan Zhang$^\ddag$, 
    ~~Weihua Peng$^\ddag$, \\
    \textbf{Yajuan Lyu}$^\ddag$,
    ~~\textbf{Zhifang Sui}$^\dag$,
    ~~\textbf{Baobao Chang}$^\dag$,
    ~~\textbf{Yong Zhu}$^\ddag$ \\
    $^\dag$MOE Key Lab of Computational Linguistics, Peking University, Beijing, China \\
    $^\ddag$Baidu Inc., Beijing, China \\
    \texttt{\{daidamai,szf,chbb\}@pku.edu.cn} \\
    \texttt{\{jiangwenbin,zhangjiyuan01,pengweihua,lvyajuan,zhuyong\}@baidu.com} \\
}

\date{}

\begin{document}

\maketitle

\begin{abstract}

Biomedical Question Answering~(BQA) has attracted increasing attention in recent years due to its promising application prospect. 
It is a challenging task because the biomedical questions are professional and usually vary widely. 
Existing question answering methods answer all questions with a homogeneous model, leading to various types of questions competing for the shared parameters, which will confuse the model decision for each single type of questions. 
In this paper, in order to alleviate the parameter competition problem, we propose a Mixture-of-Expert~(MoE) based question answering method called \ours{} that decouples the computation for different types of questions by sparse routing. 
To be specific, we split a pretrained Transformer model into bottom and top blocks. 
The bottom blocks are shared by all the examples, aiming to capture the general features. 
The top blocks are extended to an MoE version that consists of a series of independent experts, where each example is assigned to a few experts according to its underlying question type. 
\ours{} automatically learns the routing strategy in an end-to-end manner so that each expert tends to deal with the question types it is expert in. 
We evaluate \ours{} on three BQA datasets constructed based on real examinations. 
The results show that our MoE extension significantly boosts the performance of question answering models and achieves new state-of-the-art performance. 
In addition, we elaborately analyze our MoE modules to reveal how \ours{} works and find that it can automatically group the questions into human-readable clusters. 

\end{abstract}

\section{Introduction}

In recent years, Biomedical Question Answering~(BQA) has attracted increasing attention due to its promising application prospect, e.g., supporting the clinical decision for doctors, or being integrated into search engines and chatbots. 
Compared with general domain question answering, BQA is more challenging because the biomedical questions are professional and usually vary widely. 
Existing question answering methods usually answer all questions with a homogeneous model, even if different types of biomedical questions have different focuses and need different problem-solving processes. 
In this manner, different types of questions will compete for the shared model parameters, which makes a model more confused about each single type of questions and thus negatively affects the performance. 
Therefore, existing general domain question answering methods may not be the best choice for BQA. 

In this paper, we aim to alleviate the parameter competition problem via a Mixture-of-Expert~(MoE) based question answering method called \ours{} that decouples the computation for different types of questions by sparse routing. 
To be specific, we split a pretrained Transformer model into bottom and top blocks. 
The bottom blocks are shared by all the examples, aiming to capture the general features among them. 
By contrast, the top blocks are extended to an MoE version that consists of a series of independent experts. 
In the MoE-extended top blocks, each example is assigned to a few experts according to its question representation, which implies its underlying question type. 
\ours{} automatically learns the routing strategy in an end-to-end manner so that the questions will be grouped into clusters and each expert tends to answer several types of questions it is expert in. 

We evaluate \ours{} on three BQA datasets constructed based on real examinations, including MedQA~\citep{medqa}, HEAD-QA~\citep{headqa}, and NLPEC~\citep{kmqa}. 
The experimental results show that \ours{} significantly boosts the performance of question answering models and achieves new state-of-the-art performance within a tolerable computational overhead. 
In addition, we elaborately analyze our MoE modules to validate our design and reveal how \ours{} groups the questions into human-readable clusters. 

Our contributions are summarized as follows: 
(1) We point out the parameter competition problem in BQA, which limits the performance of BQA models. 
(2) We propose an MoE-based question answering method called \ours{} to alleviate the parameter competition problem by decoupling the computation for different types of questions. 
(3) We evaluate \ours{} on three BQA datasets and show that it achieves new state-of-the-art performance. We also perform quantitative and qualitative analysis on \ours{} to reveal how it works. 





\section{Multiple Choice Question Answering}
\label{sec:qa_paradigm}

In this section, we formulate the multiple choice question answering task, which is used to evaluate our method in this paper. 
In addition, we describe a typical paradigm for this task that utilizes pretrained Transformer models such as BERT~\citep{bert}. 

Given a context $C$, a question $Q$, and a set of candidate options $\mathcal{O} = \left\{ O_1, O_2, ..., O_n \right\}$, the multiple choice question answering task requires a model to select the correct option $O_a$ from the candidate set $\mathcal{O}$. 
In the typical paradigm, for each option, we first jointly encode the context $C$, the question $Q$, and the option $O_i$ to obtain an overall representation. 
Taking BERT as an example encoder, we concatenate $C$, $Q$, $O_i$, and the special tokens for BERT to form the input sequence $I = \text{``} \text{[CLS]} \; C \; Q \; \text{[SEP]} \; O_i \; \text{[SEP]} \text{''}$ and feed it into BERT to obtain the overall representation $\mathbf{p}_i$: 
\begin{align}
    H & = \operatorname{BERT}(I), \\
    \mathbf{p}_i & = \operatorname{Pooling}(H),
\end{align}
where $H$ is the hidden states computed by BERT, and $\operatorname{Pooling}(\cdot)$ pools the hidden states into a single vector $\mathbf{p}_i$ (e.g., taking the hidden state of the [CLS] token as $\mathbf{p}_i$). 
After this, we project each $\mathbf{p}_i$ into a scalar score $e_i$, and determine the predicted answer as follows: 
\begin{align}
    e_i & = \mathbf{q}^{T} \mathbf{p}_i, \\
    P(O_i | C, Q) & = \alpha_i = \frac{\operatorname{exp}(e_i)}{\sum_{j=1}^{n}{\operatorname{exp}(e_j)}} \\
    \hat{a} & = \mathop{\arg\max}\limits_{i} \alpha_i,
\end{align}
where $\mathbf{q}$ is a trainable vector, $e_i$ is the matching score for the $i$-th option, $\alpha_i$ is the predicted probability that the $i$-th option is the answer, and $\hat{a}$ is the index of the predicted answer. 

The training object of multiple choice question answering is to minimize the negative log likelihood loss: 
\begin{align}
    \mathcal{L}_{\text{task}} & = -\sum_{C, Q, \mathcal{O}}\log P(O_{a} | C, Q),
\end{align}
where $O_{a}$ is the ground truth option. 

\begin{figure}[t]
\centering
\includegraphics[width=0.98\linewidth]{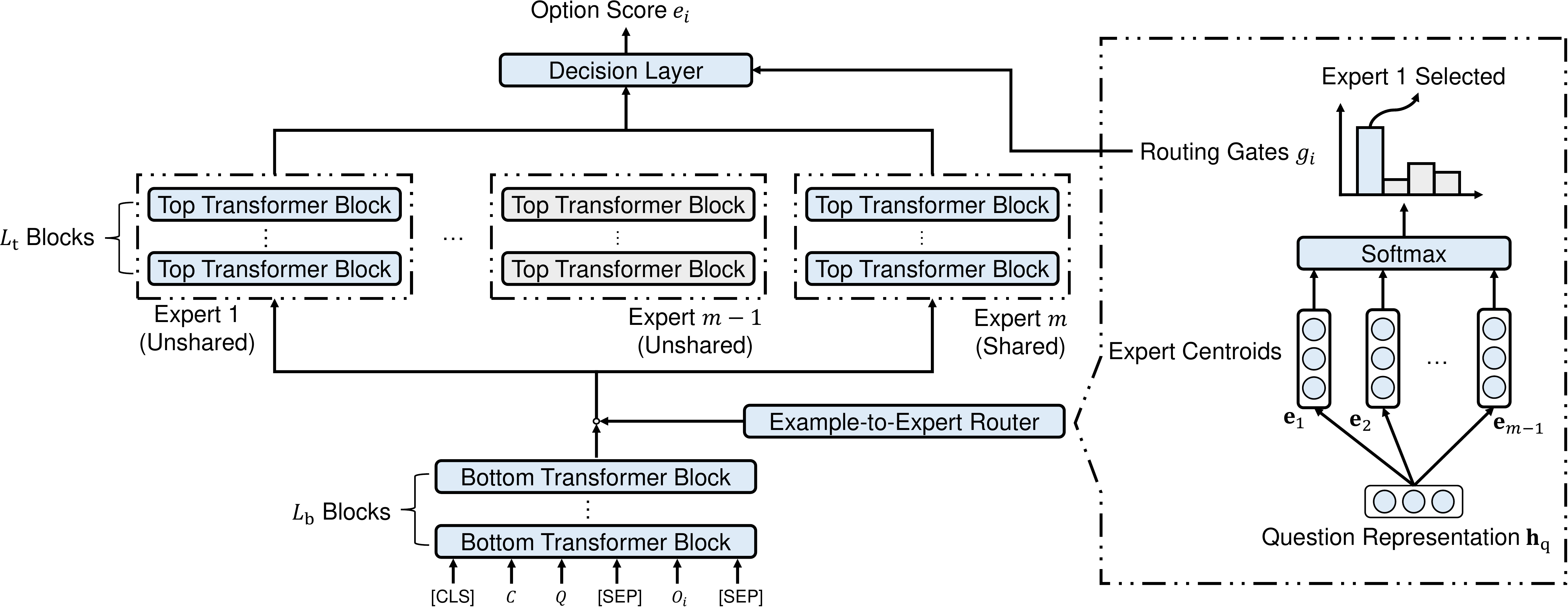}
\caption{
Illustration of \ours{}. 
The shared bottom blocks capture general features among all the examples. 
The top blocks are extended to an MoE version that consists of a shared expert and $m\text{-}1$ unshared experts. 
For each example, we assign it to the shared expert and an unshared expert according to its question representation. 
}
\label{fig:method}
\end{figure}

\section{Method: \ours{}}

Under the paradigm described in Section~\ref{sec:qa_paradigm}, \ours{} extends the pretrained Transformer model to an MoE version. 
Although our extension is applicable for all the pretrained Transformer models, we take BERT~\citep{bert} as an example backbone model to introduce our MoE extension for simplicity. 

\subsection{Overview of \ours{}}

As illustrated in Figure~\ref{fig:method}, we split a pretrained Transformer model into two parts: $L_{\text{b}}$ bottom Transformers blocks and $L_{\text{t}}$ top Transformers blocks. 
We keep the bottom blocks shared by all the examples, but extend the top blocks to an MoE version that consists of $m$ experts, including a shared one and $m\text{-}1$ unshared ones. 
Before the MoE modules, an example-to-expert router assigns each example to the shared expert and an unshared expert according to its question representation. 
In this manner, each unshared expert tends to answer a group of questions it is expert in, which alleviates the parameter competition among different types of questions. 

\subsection{Shared Bottom Blocks}

Although there exists the parameter competition among different types of questions, the examples still share some general features, e.g., the low-level features of the texts in the biomedical domain. 
Therefore, we keep the bottom $L_{\text{b}}$ Transformer blocks shared by all the examples to capture the general features and preliminarily understand the texts. 

For each input sequence $I$, the bottom blocks of our model encode it into a series of hidden states $H^{\prime}$:
\begin{equation}
    H^{\prime} = \operatorname{BERT}_{\text{bottom}}(I) = [\mathbf{h}^{L_{\text{b}}}_1; \mathbf{h}^{L_{\text{b}}}_2; ...; \mathbf{h}^{L_{\text{b}}}_T],
\end{equation}
where $\operatorname{BERT}_{\text{bottom}}$ denotes the bottom part of BERT that contains $L_{\text{b}}$ Transformer blocks, and $\mathbf{h}^{L_{\text{b}}}_t$ is the hidden state of the $t$-th token after the bottom $L_{\text{b}}$ Transformer blocks. 

\subsection{MoE-extended Top Blocks}

We extend the top $L_{\text{t}}$ Transformer blocks to an MoE version. 
To be specific, we copy the top blocks for $m$ times to produce a set of experts $\{ \operatorname{BERT}_{\text{top}}^{1}, \operatorname{BERT}_{\text{top}}^{2}, ..., \operatorname{BERT}_{\text{top}}^{m} \}$, where $\operatorname{BERT}_{\text{top}}^{m}$ is a shared expert and the other experts are unshared. 
For each example, we assign it to two experts, including the shared expert and an unshared expert. 

In order to decide which unshared expert to select, we need to compute the affinities between the question and all the unshared experts. 
Let $q_{\text{s}}$ and $q_{\text{e}}$ be the start index and the end index of the question in the input sequence $I$, respectively. 
We first compute the question representation $\mathbf{h}_{\text{q}}$ by mean pooling the token representations of the question in $H^{\prime}$:
\begin{equation}
    \mathbf{h}_{\text{q}} = \operatorname{MeanPooling}(\mathbf{h}^{L_{\text{b}}}_{q_{\text{s}}:q_{\text{e}}}). 
\end{equation}
For each unshared expert $\operatorname{BERT}_{\text{top}}^{i}$, we define a trainable centroid vector $\mathbf{e}_i$ for it. 
Then, the affinity between the question and the $i$-th expert is computed by
\begin{equation}
    s_i = \mathbf{e}_i^{\top} \mathbf{h}_{q}. 
\end{equation}
Given the affinity score $s_i$, we greedily select the $t$-th expert with the highest affinity as the target expert: 
\begin{equation}
    t = \mathop{\arg\max}\limits_{i} s_i. 
\end{equation}
Finally, we compute the output of the MoE top blocks as
\begin{align}
    & \quad g_i = \frac{\operatorname{exp}(s_i)}{\sum_{j=1}^{m-1}{\operatorname{exp}(s_j)}}, \label{equ:gate} \\
    H = (1 - g_t) & \operatorname{BERT}_{\text{top}}^{m}(H^{\prime}) + g_t \cdot \operatorname{BERT}_{\text{top}}^{t}(H^{\prime}),
\end{align}
where $g_t$ is a softmax gate that controls how much the $t$-th expert will be used. 
Considering the gate $g_t$, if the selected unshared expert $\operatorname{BERT}_{\text{top}}^{t}$ is beneficial for answering the question, optimizing the training objective $\mathcal{L}_{\text{task}}$ will urge the gate to be larger; otherwise, the model will tend to use the expert sparely by suppressing the gate. 
Therefore, the gate $g_t$ can encourage similar questions to be assigned to the same unshared expert that is beneficial to them, i.e., automatically grouping the examples into clusters according to their underlying question types. 

\paragraph{Balance Loss}
In ideal conditions, we expect each expert to have a relatively balanced example load to guarantee the parameter utilization. 
Otherwise, some experts will degrade into ineffective ones since they are seldom activated. 
Therefore, we adopt a differentiable balance loss~\citep{switch,gshard} to avoid imbalanced expert loads. 
Let $c_i$ denotes the number of examples that the $i$-th expert has been assigned in the training history, the balance loss $\mathcal{L}_{\text{bal}}$ is computed as follows: 
\begin{equation}
    \mathcal{L}_{\text{bal}} = (m - 1) \sum_{i=1}^{m-1}{\frac{c_i}{\sum_{j=1}^{m-1}{c_j}} g_i },
\end{equation}
Intuitively, if an expert is overloaded, the weight $\frac{c_i}{\sum_{j=1}^{m-1}{c_j}}$ will be higher than the average, and thus the balance loss tends to decrease the affinities related to the expert to drop some examples. 
Otherwise, if an expert is unoccupied, the balance loss will increase its affinities to capture more examples. 
The balance loss is minimized when the experts have absolutely balanced loads. 

\subsection{Training Objective}

The training objective of our method is almost the same as in the typical paradigm, except that we add the balance loss: 
\begin{equation}
    \mathcal{L}_{\text{train}} = \mathcal{L}_{\text{task}} + \beta \mathcal{L}_{\text{bal}}, 
\end{equation}
where the balance factor $\beta$ is a hyper-parameter, which should be large enough to guarantee the balanced loads and meanwhile not too significant to overwhelm the effect of the primary task loss $\mathcal{L}_{\text{task}}$.

\section{Experiments}

\begin{table}[t]
\centering
\setlength{\tabcolsep}{15pt}
\begin{tabular}{l c c c c}
\toprule
\textbf{Datasets} & \textbf{Training} & \textbf{Validation} & \textbf{Test} & \textbf{Total} \\
\midrule
MedQA & 10,178 & 1,272 & 1,273 & 12,723 \\
HEAD-QA & ~~2,657 & 1,366 & 2,742 & ~~6,765 \\
NLPEC & 18,703 & 2,500 & ~~~547 & 21,750 \\
\bottomrule
\end{tabular}
\caption{
Official data splits of three BQA datasets used in this paper. 
}
\label{tab:dataset}
\end{table}

\subsection{Datasets}

We evaluate \ours{} on three multiple choice BQA datasets, including two English datasets, MedQA~\citep{medqa}, HEAD-QA~\citep{headqa}, and a Chinese dataset, NLPEC~\citep{kmqa}. 

\textbf{MedQA} is extracted from the National Medical Board Examinations in the USA, Mainland China, and Taiwan. 
The full dataset covers three languages, including English, simplified Chinese, and traditional Chinese. 
In this paper, we use only the English subset, which contains $12,723$ examples in total. 
Each example in the dataset contains a question and several options with the correct one annotated. 
\textbf{HEAD-QA} is created from real examinations spanning from 2013 to 2017 that are organized by the Spanish government and used for applying for specialization positions in public healthcare areas. 
The original dataset is in Spanish, but it also has an official English version, which is used in this paper. 
HEAD-QA contains $6,765$ examples in total, where each example has a question and several options with the correct one annotated. 
\textbf{NLPEC} is a Chinese dataset, constructed based on the National Licensed Pharmacist Examination in China. 
It contains $21,750$ examples in total, where each example in NLPEC has a question, several options with the correct one annotated, and relevant evidences extracted from the official exam guide book. 
The official data splits of these three datasets are shown in Table~\ref{tab:dataset}. 

\begin{table}[t]
\centering
\setlength{\tabcolsep}{10pt}
\begin{tabular}{l c c}
\toprule
\textbf{Methods} & \textbf{Valid Accuracy} & \textbf{Test Accuracy} \\
\midrule
ClinicalBERT-base & 33.7 & 32.4 \\
BioBERT-base & 34.3 & 34.1 \\
BERT-base & 33.9 & 34.3 \\
BioRoBERTa-base & 35.1 & 36.1 \\
RoBERTa-large & 35.2 & 35.0 \\
BioBERT-large & 36.1 & 36.7 \\
PubMedBERT & - & 35.1 \\
PubMedBERT + MoP$^{\clubsuit}$ & - & 38.0 \\
\midrule
PubMedBERT + \ours{} & \textbf{39.9} & \textbf{41.6} \\
\bottomrule
\end{tabular}
\caption{
Accuracy on MedQA. 
The best performance is marked in \textbf{bold}. 
${\clubsuit}$ denotes the existing SOTA method. 
\ours{} achieves new SOTA performance. 
}
\label{tab:medqa}
\end{table}

\subsection{Experimental Setup}

We conduct experiments based on the HuggingFace transformers library\footnote{https://github.com/huggingface/transformers}.
All experiments are conducted on NVIDIA V100 GPUs with 32 GB memory. 
We evaluate the model on the validation set after each epoch and use the best checkpoint that achieves the highest validation accuracy to obtain the test accuracy on the test set. 
Due to the space limit, we describe the key settings for each dataset separately as follows, and put more details in Appendix A. 

\paragraph{MedQA}
For MedQA, we use PubMedBERT~\citep{pubmedbert} as the pretrained backbone model, which shares the same architecture as BERT-base~\citep{bert}, but is pretrained from scratch on biomedical domain corpora. 
We use AdamW~\citep{adamw} with $\beta_1=0.9$ and $\beta_2=0.999$ as the optimizer, and set the learning rate to 3e-5. 
We set the batch size to 16 and train the model for 5 epochs. 
We set $L_{b}$ to 10 and $L_{t}$ to 2. 
The number of experts is set to 5 and the balance factor $\beta$ is set to $0.01$. 

\paragraph{HEAD-QA}
For HEAD-QA, we also use PubMedBERT as the backbone model. 
Since the questions in HEAD-QA do not contain enough contextual information, we follow \citep{headqa} to use the DrQA document retriever~\citep{drqa} to retrieve a relevant document from Wikipedia as the context. 
We use AdamW with $\beta_1=0.9$ and $\beta_2=0.999$ as the optimizer, and set the learning rate to 5e-5. 
We set the batch size to 8 and train the model for 2 epochs. 
We set $L_{b}$ to 10 and $L_{t}$ to 2. 
The number of experts is set to 3 and the balance factor $\beta$ is set to $0.003$. 

\paragraph{NLPEC}
For NLPEC, we use RoBERTa-large~\citep{roberta} as the backbone model. 
We use AdamW with $\beta_1=0.9$ and $\beta_2=0.999$ as the optimizer, and set the learning rate to 2e-5. 
We set the batch size to 16 and train the model for 35 epochs. 
We set $L_{b}$ to 20 and $L_{t}$ to 4. 
The number of experts is set to 5 and the balance factor $\beta$ is set to $0.001$. 

\begin{table}[t]
\centering
\setlength{\tabcolsep}{10pt}
\begin{tabular}{l c c}
\toprule
\textbf{Methods} & \textbf{Valid Accuracy} & \textbf{Test Accuracy} \\
\midrule
BiDAF & - & 30.3 \\
TFIDF-IR & - & 37.2 \\
IR + BERT & - & 35.0 \\
IR + BioBERT & - & 36.4 \\
Multi-step Reasoner & - & 42.9 \\
PubMedBERT & 43.5 & 45.6 \\
\midrule
MurKe (30 documents)$^{\clubsuit}$ & - & \textbf{46.7} \\
MurKe (10 documents) & - & $<$ 40.0~~~~ \\
\midrule
PubMedBERT + \ours{} (1 document) & \textbf{44.3} & \textbf{46.7} \\
\bottomrule
\end{tabular}
\caption{
Accuracy on HEAD-QA. 
MurKe needs to leverage 30 supporting documents to achieve the SOTA accuracy, but \ours{} uses only one supporting document to achieve the same SOTA accuracy. 
}
\label{tab:headqa}
\end{table}

\begin{table}[t]
\centering
\setlength{\tabcolsep}{19pt}
\begin{tabular}{l c c}
\toprule
\textbf{Methods} & \textbf{Valid Accuracy} & \textbf{Test Accuracy} \\
\midrule
BiDAF & 52.7 & 43.6 \\
BERT-base & 64.2 & 52.2 \\
ERNIE & 64.7 & 53.4 \\
RoBERTa-large & 70.8 & 57.9 \\
BERT-base + KMQA & 67.9 & 57.1 \\
RoBERTa-large + KMQA$^{\clubsuit}$ & 71.1 & 61.8 \\
\midrule
RoBERTa-large + \ours{} & \textbf{72.8} & \textbf{62.2} \\
\bottomrule
\end{tabular}
\caption{
Accuracy on NLPEC. 
\ours{} achieves new SOTA performance. 
}
\label{tab:nlpec}
\end{table}

\begin{table}[t]
\centering
\setlength{\tabcolsep}{8pt}
\begin{tabular}{l c c c c}
\toprule
\multirow{2}{*}{\textbf{Datasets}} & \multicolumn{2}{c}{\textbf{Training}} &
\multicolumn{2}{c}{\textbf{Inference}} \\
\cmidrule(lr){2-3}\cmidrule(lr){4-5}
 & \textbf{Dense Backbone} & \textbf{+\ours{}} & \textbf{Dense Backbone} & \textbf{+\ours{}} \\
\midrule
MedQA & 39.3 & 30.0 & 113.3 & 91.3 \\
HEAD-QA & 15.5 & 12.8 & ~~60.5 & 50.3 \\
NLPEC & ~~6.4 & ~~5.5 & ~~19.2 & 16.5 \\
\bottomrule
\end{tabular}
\caption{
Number of examples that the dense backbone model and \ours{} can process per second on three datasets. 
The computational overhead introduced by our MoE extension is tolerable. 
}
\label{tab:speed}
\end{table}

\subsection{Results}

\paragraph{MedQA}
On MedQA, we compare \ours{} with ClinicalBERT~\citep{clinicalbert}, BioBERT~\citep{biobert}, BERT~\citep{bert}, BioRoBERTa~\citep{bioroberta}, RoBERTa~\citep{roberta}, and PubMedBERT + MoP~\citep{pubmedbert,mop}. 
The results on MedQA are shown in Table~\ref{tab:medqa}. 
From the table, we find that \ours{} significantly outperforms all the previous BQA methods and achieves new state-of-the-art performance. 

\paragraph{HEAD-QA}
On HEAD-QA, we compare \ours{} with BiDAF~\citep{bidaf}, TFIDF-IR~\citep{headqa}, IR + BERT~\citep{murke}, IR + BioBERT~\citep{murke}, Multi-step Reasoner~\citep{murke}, PubMedBERT~\citep{pubmedbert}, and MurKe~\citep{murke}. 
As shown in Table~\ref{tab:headqa}, \ours{} improves the test accuracy of PubMedBERT by 1.1 and achieves state-of-the-art performance. 
The test accuracy of MurKe is the same as ours, but MurKe needs 30 supporting documents for reference while we need only one. 
If the number of supporting documents for MurKe is limited to 10, its test accuracy will fall below 40.0. 
Therefore, from the aspect of the modeling ability, \ours{} is stronger than all the existing BQA methods. 

\paragraph{NLPEC}
On NLPEC, we compare \ours{} with BiDAF, BERT, RoBERTa, ERNIE~\citep{ernie_baidu}, and KMQA~\citep{kmqa}. 
We show the results in Table~\ref{tab:nlpec} and observe that \ours{} improves the test accuracy of RoBERTa-large by 4.3 and achieves new state-of-the-art performance. 

\paragraph{Computational Efficiency}
In Table~\ref{tab:speed}, we report the number of examples that the dense backbone model and \ours{} can process per second on three datasets. 
Compared to the dense model, \ours{} is slightly slower. 
However, considering the significant performance improvement, the computational overhead introduced by our MoE extension is tolerable. 

\section{Analysis and Discussion}

\begin{table}[t]
\centering
\setlength{\tabcolsep}{22pt}
\begin{tabular}{l l l}
\toprule
\textbf{Methods} & \textbf{Valid Accuracy} & \textbf{Test Accuracy} \\
\midrule
\ours{} & 39.9 & 41.6 \\
w/o Balance Loss & 38.6~($1.3\downarrow$) & 40.1~($1.5\downarrow$) \\
w/o Gate Signal & 39.2~($0.7\downarrow$) & 39.7~($1.9\downarrow$) \\
w/o Shared Expert & 38.0~($1.9\downarrow$) & 39.7~($1.9\downarrow$) \\
w/o MoE Extension & 38.3~($1.6\downarrow$) & 38.7~($2.9\downarrow$) \\
Routing by [CLS] & 37.6~($2.3\downarrow$) & 38.2~($3.4\downarrow$) \\
\bottomrule
\end{tabular}
\caption{
Ablation studies of \ours{}. 
}
\label{tab:ablation}
\end{table}

\subsection{Ablation Studies}
We show the ablation studies of \ours{} on the MedQA dataset in Table~\ref{tab:ablation} to validate the effectiveness of our design. 
Firstly, if we remove the balance loss, unbalanced expert loads will affect the parameter utilization and thus decrease the test accuracy by 1.5. 
Secondly, if we replace the gate $g_t$~(see Equation~\ref{equ:gate}) with a constant $0.5$, no useful signal will be propagated from the task loss back to the routing strategy. 
As a result, the model will produce worse question clusters, leading to a decrease of 1.9 in the test accuracy. 
Thirdly, if we remove the shared expert and all the experts are unshared, the test accuracy will drop by 1.9. 
Finally, if we disable the whole MoE extension, the test accuracy will significantly drop by 2.9. 
In addition, we investigate the effects of the routing features. 
By default, we compute a question representation as the routing feature. 
If we use the representation of the [CLS] token for routing, the test accuracy will drop by 3.4. 
The result indicates that the question representation is a better routing feature in \ours{} than the whole example representation. 

\begin{figure}[t]
\centering
\includegraphics[width=0.99\linewidth]{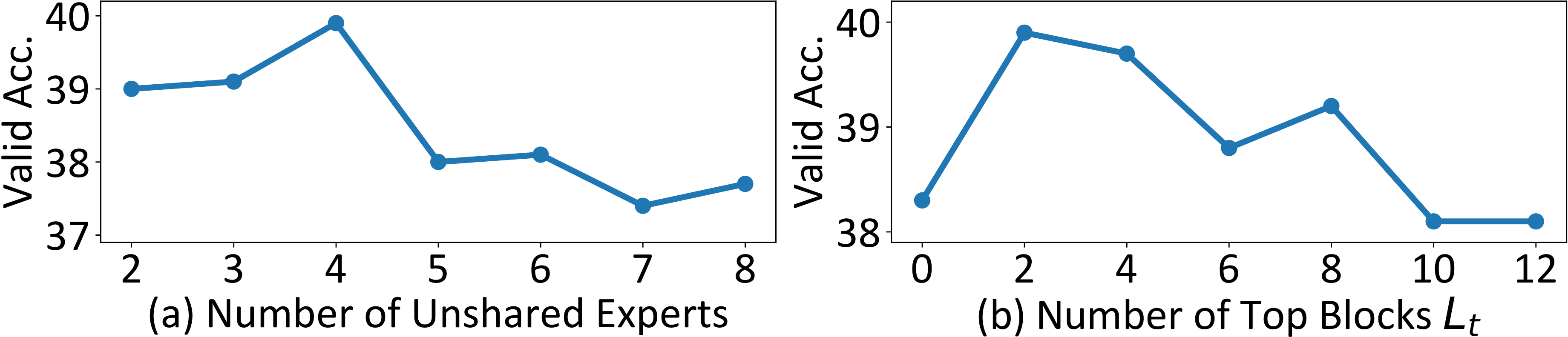}
\caption{
Performance of different numbers of unshared experts (Subplot a) and different numbers of top blocks $L_{t}$ (Subplot b). 
}
\label{fig:arch}
\end{figure}


\subsection{Investigation of MoE Architectures}

Based on the MedQA dataset, we investigate the performance of different MoE architectures, including different numbers of unshared experts and different splitting ratios between the bottom and top Transformers blocks. 

The performance of different numbers of unshared experts is plotted in Figure~\ref{fig:arch}(a). 
We find that adding experts can improve the performance when the number of unshared experts is smaller than 4. 
However, too many experts will harm the performance instead, since the number of experts affects the granularity of the question clusters. 
If the number of experts is too large, each question cluster will be too small, which will damage the generalization performance of the model. 

The performance of different splitting ratios between $L_{\text{b}}$ and $L_{\text{t}}$ is plotted in Figure~\ref{fig:arch}(b). 
When $L_{\text{t}}$ is 0, the model is equivalent to a dense model that does not have MoE modules. 
Based on it, extending 2 top blocks to the MoE version can significantly boost the performance. 
However, continuing to increase $L_{\text{t}}$ will lead to poorer performance. 
The results prove that the shared bottom blocks in \ours{}, which capture the general features among all the examples, are also indispensable and we should allocate a proper proportion to $L_{\text{b}}$. 

\begin{figure}[t]
\centering
\includegraphics[width=0.6\linewidth]{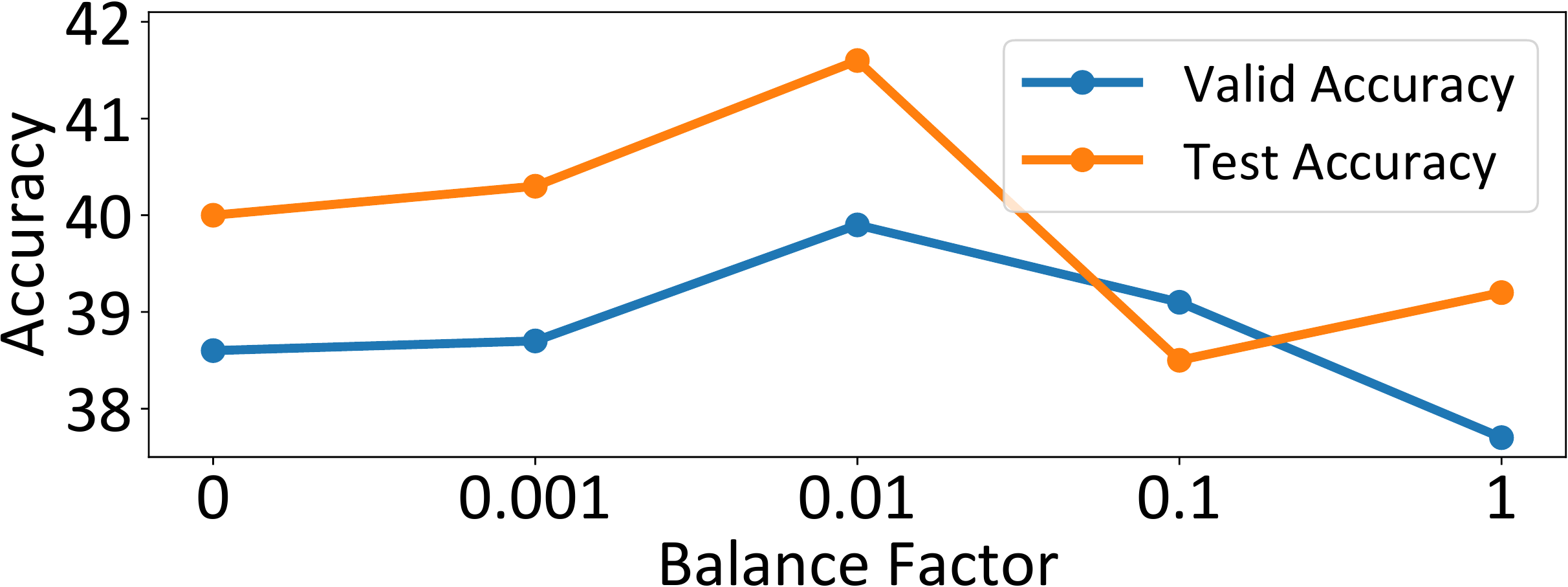}
\caption{
Effects of different balance factors. 
Within a tolerable range, a larger balance factor can improve the performance by making the expert loads more balanced. 
However, too large balance factors will overwhelm the primary loss and damage the performance. 
}
\label{fig:balance}
\end{figure}

\subsection{Effects of Balance Factor}

Figure~\ref{fig:balance} shows the effects of different balance factors on the MedQA dataset. 
When the balance factor is 0, the balance loss does not take effect. 
As we gradually increase the balance factor, we have the following observations. 
Within a tolerable range, a larger balance factor can improve the performance by making the expert loads more balanced, and we achieve the best performance when the balance factor is 0.01. 
However, if the balance factor is too large, the balance loss will overwhelm the primary loss and thus damage the performance. 
Therefore, the performance with the balance factor of 1 is even worse than that when not applying the balance loss. 

\subsection{Diversity of Experts}

\begin{figure}[t]
\centering
\includegraphics[width=0.6\linewidth]{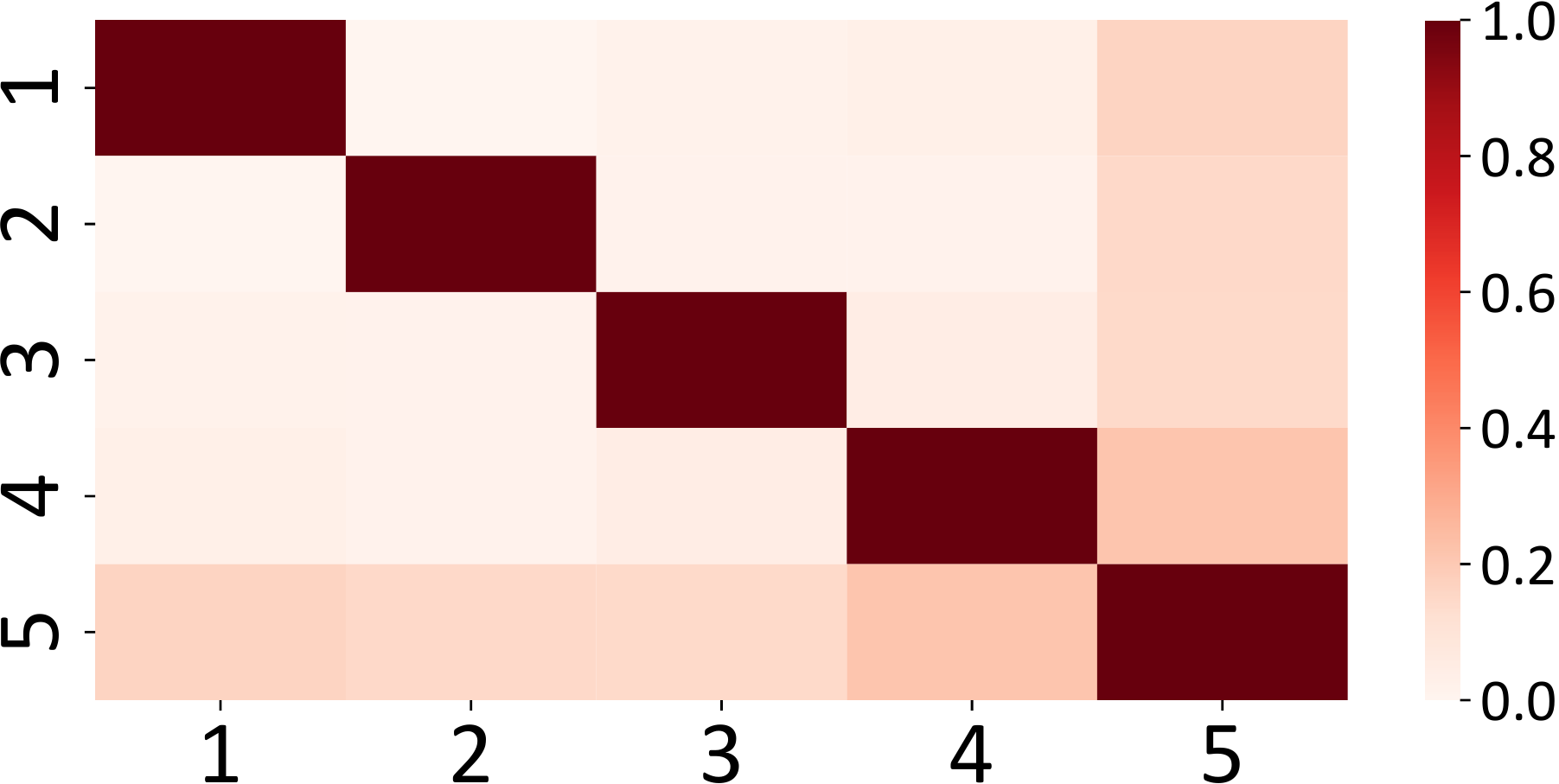}
\caption{
Normalized similarities between experts. 
Each unshared expert~(Experts 1-4) has a moderate similarity with the shared expert~(Expert 5) and relatively low similarities with the other unshared experts. 
}
\label{fig:heatmap}
\end{figure}

In order to demonstrate the diversity of different experts, we show the similarity between each pair of experts in Figure~\ref{fig:heatmap}. 
To be specific, for the best model trained on MedQA, we flatten all the parameters in each expert, calculate the cosine similarity between each pair of experts, and normalize the similarities into the range of $[0, 1]$. 
From the figure, we observe that each expert has the highest similarity with itself and a moderate similarity with the shared expert~(Expert 5). 
Meanwhile, each unshared expert has relatively low similarities with the other unshared experts. 
The results prove that the unshared experts have considerable diversity and the shared expert tends to understand relatively common information. 

\begin{table}[t]
\centering
\footnotesize
\setlength{\tabcolsep}{12pt}
\begin{tabular}{@{}c l l@{}}
\toprule
\textbf{Expert ID} & \textbf{Questions with Highest Affinities} & \textbf{Common Topic} \\
\midrule
\multirow{3}{*}{1} & What is the most likely organism responsible? & \multirow{3}{*}{Pathogenesis} \\
 & Which of the following is the most likely pathogen? & \\
 & Which of the following is the most probable cause of his complaints? & \\
\midrule
\multirow{3}{*}{2} & Which of the following is the best next step in management? & \multirow{3}{*}{Treatments} \\
 & Which of the following is the most appropriate next step in management? & \\
 & Which of the following is the initial treatment of choice for the patient? & \\
\midrule
\multirow{3}{*}{3} & Further evaluation is most likely to show which of the following? & \multirow{3}{*}{Deduction} \\
 & This drug is most likely to result in which of the following? & \\
 & Microscopic examination of the mass will most likely show which of the following? & \\
\midrule
\multirow{3}{*}{4} & What disease is the child suffering from? & \multirow{3}{*}{Diagnosis} \\
 & What vitamin deficiency is this woman most likely suffering from? & \\
 & What is the pathophysiology of this patient’s condition? & \\
\bottomrule
\end{tabular}
\caption{
Three representative questions in the validation set with the highest affinities to each expert. 
High-affinity questions assigned to the same expert tend to ask about a common topic, which is summarized in the third column. 
}
\label{tab:case_study}
\end{table}

\subsection{Case Study}

We present a case study of \ours{} in Table~\ref{tab:case_study} to reveal the behavior of our MoE modules intuitively. 
For each expert, we compute its affinities with all the questions in the MedQA validation set. 
Then, we present three representative questions among the top-5 questions with the highest affinities. 
From Table~\ref{tab:case_study}, we observe that the questions assigned to the same expert tend to ask about a common topic. 
For example, 
(1) the questions assigned to Expert 1 require identifying the pathogenesis, 
(2) the questions assigned to Expert 2 ask for suggestions on the treatments, 
(3) the questions assigned to Expert 3 need to deduce an expected result, 
(4) and the questions assigned to Expert 4 aim to diagnose the disease. 
These cases prove that \ours{} can automatically detect the underlying types of questions and group the questions into clusters. 
By dealing with each cluster of questions separately, \ours{} achieves better performance than the dense model. 

\section{Related Work}

\paragraph{Biomedical Question Answering}
In recent years, many efforts in constructing BQA datasets are paid to facilitate the BQA research. 
BioASQ~\citep{bioasq} is a cornerstone challenge that spans several years and has various types of task types. 
emrQA~\citep{emrqa} is a large-scale reading comprehension dataset generated by an automatic algorithm. 
Considering that the question-answer pairs in real examinations usually have a considerable scale and high quality, MedQA~\citep{medqa}, HEAD-QA~\citep{headqa}, and NLPEC~\citep{kmqa} are constructed based on medical examinations in different countries and districts. 
These examination-based datasets can reflect the ability of a model to solve problems in real-world scenarios, so we focus on them to validate the effectiveness of our method. 

\paragraph{Mixture of Experts}
\citet{ori_moe1} first propose the MoE technique to compute different examples with independent experts. 
\citet{moe} first apply MoE to build large-scale LSTM language models. 
Recently, MoE-version feed-forward networks has been designed for MoE-based Transformers. 
GShard~\citep{gshard}, Switch Transformer~\citep{switch}, and BASE Layer~\citep{base} dynamically learn to route each input token to experts. 
Hash Layer~\citep{hash} propose to use a pre-designed token-level hash table as the routing strategy. 
StableMoE~\citep{stable_moe} first learns to route and then fixes the learned routing strategy for more stable training. 
Our work applies MoE to BQA at an example granularity, aiming to decouple the computation for different types of questions. 

\section{Conclusions}

In this paper, we point out the parameter competition problem among different types of questions in BQA. 
In order to alleviate this problem, we propose a Mixture-of-Expert~(MoE) based question answering method called \ours{} that decouples the computation for different types of questions by sparse routing. 
We evaluate \ours{} on three BQA datasets constructed based on real examinations and achieve new state-of-the-art performance. 
In addition, we elaborately analyze our MoE modules to reveal how \ours{} works. 
Note that our MoE extension is neither limited to Transformer models nor limited to question answering tasks, so it has the potential to be generalized to other models and tasks in the future. 
Also, although we conduct experiments on only BQA, we encourage to try the idea of MoE extension on other question answering tasks as long as they also need to decouple the computation for different types of questions. 

\bibliographystyle{coling}
\bibliography{coling2020}

\clearpage
\appendix

\section*{Appendix}

\section{Details of Hyper-parameters}

We present the details of our hyper-parameters in this section, including the search range and the finally selected value. 
We use the grid search to find the best configuration that achieves the highest accuracy. 
For our reproduced baselines, we use the same search range for the hyper-parameters that are valid for them (i.e., all the hyper-parameters except for those related to MoE modules). 
The hyper-parameters on MedQA, HEAD-QA, and NLPEC are presented in Table~\ref{tab:hyper_medqa},  Table~\ref{tab:hyper_headqa}, and  Table~\ref{tab:hyper_nlpec}, respectively. 

\begin{table}[ht]
\centering
\footnotesize
\setlength{\tabcolsep}{10pt}
\begin{tabular}{l c c}
\toprule
\textbf{Hyper-parameters} & \textbf{Search Range} & \textbf{Selected Value} \\
\midrule
Pretrained Backbone Model & - & PubMedBERT \\
\midrule
Number of Unshared Experts & \{2, 4\} & 4 \\
Number of Shared Experts & - & 1 \\
$L_{t}$ & \{2, 4\} & 2 \\
$L_{b}$ & - & 10 \\
Balance Factor & \{0.001, 0.003, 0.01\} & 0.01 \\
\midrule
Optimizer & - & AdamW \\
$\beta_1$ & - & 0.9 \\
$\beta_2$ & - & 0.999 \\
Weight Decay & - & 0 \\
Maximum Learning Rate & \{2e-5, 3e-5, 5e-5\} & 3e-5 \\
Learning Rate Scheduler & - & Linear Decay \\
\midrule
Training Epochs & \{2, 3, 5\} & 5 \\
Warm-up Ratio & - & 0.1 \\
\midrule
Batch Size & - & 8 \\
Number of GPUs & - & 1 \\
Gradient Accumulation & \{1, 2\} & 2 \\
Max Length & - & 512 \\
FP16 & - & True \\
\bottomrule
\end{tabular}
\caption{
The search range and the finally selected value for the hyper-parameters in \ours{} on MedQA. 
- in the second column means that we directly use an empirical value without searching (e.g., $\beta_1$ and $\beta_2$), or the value can be directly computed when another hyper-parameter is assigned (e.g., $L_{b}$). 
}
\label{tab:hyper_medqa}
\end{table}

\begin{table}[ht]
\centering
\footnotesize
\setlength{\tabcolsep}{10pt}
\begin{tabular}{l c c}
\toprule
\textbf{Hyper-parameters} & \textbf{Search Range} & \textbf{Selected Value} \\
\midrule
Pretrained Backbone Model & - & PubMedBERT \\
\midrule
Number of Unshared Experts & \{2, 4\} & 2 \\
Number of Shared Experts & - & 1 \\
$L_{t}$ & \{2, 4\} & 2 \\
$L_{b}$ & - & 10 \\
Balance Factor & \{0.001, 0.003, 0.01\} & 0.003 \\
\midrule
Optimizer & - & AdamW \\
$\beta_1$ & - & 0.9 \\
$\beta_2$ & - & 0.999 \\
Weight Decay & - & 0 \\
Maximum Learning Rate & \{2e-5, 3e-5, 5e-5\} & 5e-5 \\
Learning Rate Scheduler & - & Linear Decay \\
\midrule
Training Epochs & \{2, 3, 5\} & 2 \\
Warm-up Ratio & - & 0.1 \\
\midrule
Batch Size & - & 8 \\
Number of GPUs & - & 1 \\
Gradient Accumulation & \{1, 2\} & 1 \\
Max Length & - & 512 \\
FP16 & - & True \\
\bottomrule
\end{tabular}
\caption{
The search range and the finally selected value for the hyper-parameters in \ours{} on HEAD-QA. 
}
\label{tab:hyper_headqa}
\end{table}

\begin{table}[ht]
\centering
\footnotesize
\setlength{\tabcolsep}{10pt}
\begin{tabular}{l c c}
\toprule
\textbf{Hyper-parameters} & \textbf{Search Range} & \textbf{Selected Value} \\
\midrule
Pretrained Backbone Model & - & RoBERTa-large \\
\midrule
Number of Unshared Experts & \{2, 4\} & 4 \\
Number of Shared Experts & - & 1 \\
$L_{t}$ & - & 4 \\
$L_{b}$ & - & 20 \\
Balance Factor & \{0.001, 0.003, 0.01\} & 0.001 \\
\midrule
Optimizer & - & AdamW \\
$\beta_1$ & - & 0.9 \\
$\beta_2$ & - & 0.999 \\
Weight Decay & - & 0 \\
Maximum Learning Rate & \{2e-5, 3e-5, 5e-5\} & 2e-5 \\
Learning Rate Scheduler & - & Linear Decay \\
\midrule
Training Epochs & - & 35 \\
Warm-up Ratio & - & 0.1 \\
\midrule
Batch Size & - & 8 \\
Number of GPUs & - & 1 \\
Gradient Accumulation & - & 2 \\
Max Length & - & 512 \\
FP16 & - & True \\
\bottomrule
\end{tabular}
\caption{
The search range and the finally selected value for the hyper-parameters in \ours{} on NLPEC. 
}
\label{tab:hyper_nlpec}
\end{table}

\end{document}